\documentclass[letterpaper, 10 pt, conference]{ieeeconf} 
\IEEEoverridecommandlockouts % This command is only needed if you want \thanks command
\usepackage{float}
\usepackage{graphics} % for pdf, bitmapped graphics files
\usepackage{epsfig} % for postscript graphics files
\usepackage{times} % assumes new font selection scheme installed
\usepackage{amsmath} % assumes amsmath package installed
\usepackage{amssymb}  % assumes amsmath package installed
\usepackage[pagebackref,breaklinks]{hyperref} % dont color the hyperlinks
\usepackage{lipsum}
\usepackage{xcolor}
\usepackage{gensymb}
\usepackage{multirow}

\usepackage{cite}
\newcommand{\orcid}[1]{\href{https://orcid.org/#1}{\includegraphics[width=10pt]{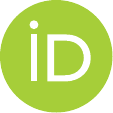}}}

\title{\LARGE \bf
FloPE: Flower Pose Estimation for Precision Pollination
}

\author{Rashik Shrestha$^{1}$ \orcid{0009-0007-7939-2623}, Madhav Rijal \orcid{0000-0001-9400-0877}, Trevor Smith \orcid{0000-0002-8921-9281}, Yu Gu \orcid{0000-0003-3165-3269}\vspace{2mm}\\
\textit{West Virginia University}\\
Morgantown, USA
\thanks{$^{1}$Rashik Shrestha is the corresponding author \href{mailto:rs00140@mix.wvu.edu}{rs00140@mix.wvu.edu}}
\thanks{This study was supported in part by USDA NIFA Award 2022-67021-36124 "Collaborative Research: NRI: StickBug - an Effective Co-Robot for Precision Pollination," the National Science Foundation Graduate Research Fellowship Award \#2136524, and by the NASA West Virginia Space Grant Consortium, Grant \#80NSSC20M0055.}
}

\begin{document}
\maketitle
\thispagestyle{empty}
\pagestyle{empty}

\begin{abstract}
This study presents Flower Pose Estimation (FloPE), a real-time flower pose estimation framework for computationally constrained robotic pollination systems. Robotic pollination has been proposed to supplement natural pollination to ensure global food security due to the decreased population of natural pollinators. However, flower pose estimation for pollination is challenging due to natural variability, flower clusters, and high accuracy demands due to the flowers' fragility when pollinating. This method leverages 3D Gaussian Splatting to generate photorealistic synthetic datasets with precise pose annotations, enabling effective knowledge distillation from a high-capacity teacher model to a lightweight student model for efficient inference. The approach was evaluated on both single and multi-arm robotic platforms, achieving a mean pose estimation error of 0.6 cm and 19.14 degrees within a low computational cost. Our experiments validate the effectiveness of FloPE, achieving up to 78.75\% pollination success rate and outperforming prior robotic pollination techniques. 

\end{abstract}

\section{INTRODUCTION}

The decline in population and diversity of natural pollinators and a growing human population have caused widespread concern for food scarcity \cite{potts2016assessment}. This has led agriculturists to seek new technologies to improve the yield and efficiency of their crops. One proposed method of improving crop yield is the utilization of robotics for pollination to enhance fertilization \cite{smith2024design}. With the help of robots, farmers can supplement natural pollination and gain access to many potential benefits. For example, robotic pollinators can gather information about crop health with sub-plant level precision \cite{weiss2011plant,cheein2011optimized,pena2013weed,mueller2017robotanist,gallmann2022flower}, allowing growers to efficiently utilize agricultural inputs (e.g., water and fertilizer), reducing waste. Also, unlike natural pollinators, robots can be configured to specific artificial environments, such as greenhouses, and accomplish other tasks in the plant growth cycle: phenotyping \cite{das2022flowerphenonet}, pruning \cite{liu2012image, schneider20233d}, and harvesting \cite{li2024ahppebot,jun2021harvest}. 

\begin{figure}
    \centering
    \includegraphics[width=\linewidth]{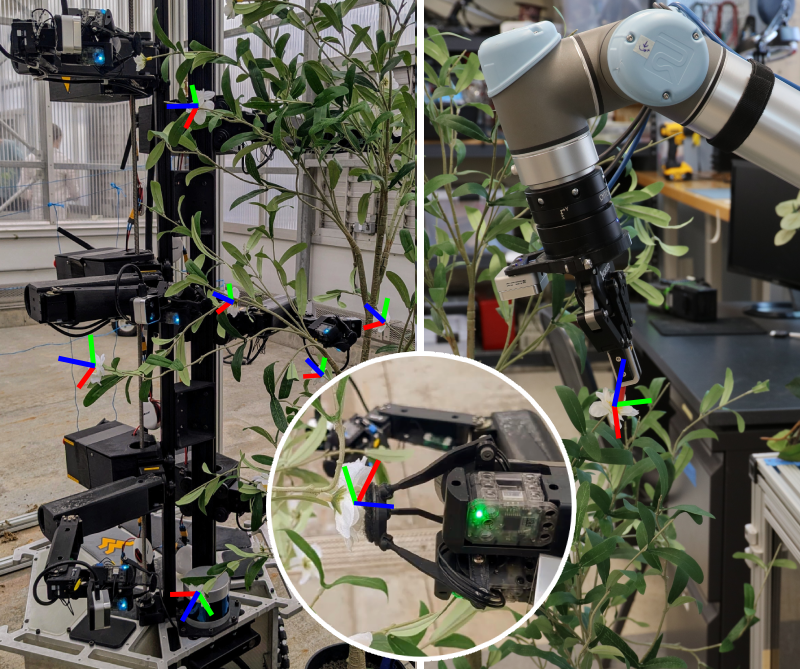}
    \caption{\textbf{Stickbug~\cite{smith2024design} (left) and UR5 (right) performing precision pollination task using estimated flower poses.} {\color{blue}Blue} Z-axis indicates the flower's facing direction. {\color{red}Red} and {\color{green}Green} axes represent XY plane, not necessarily the direction of X and Y axes, as our experimental flower has radial symmetry about Z-axis.}
    \label{fig:teaser}
    \vspace{-4mm}
\end{figure}

Agricultural tasks are inherently orientation-specific—requiring flowers to be pollinated at the pistil \cite{smith2024design}, fruit to be harvested at the stem \cite{li2024ahppebot,jun2021harvest}, and plant nodes (i.e., where leaves and flowers are attached to the stem) to be pruned orthogonally to the branch \cite{liu2012image, schneider20233d}. However, orientation estimation has often lagged behind due to the inherently complex nature of the problem and the computational limitations of end devices \cite{smith2024design}. To address this gap, this work focuses on developing 3D pose estimation of flowers with low computational requirements to support agricultural robotics.

3D pose estimation of flowers faces several challenges. First, plants' high variability and complex natural growth patterns make reliable detection difficult. Additionally, flowers often grow in dense clusters, complicating individual detection. The delicate nature of flowers also demands high accuracy to avoid damage during manipulation. Moreover, plants' flexible and deformable structures complicate the pre-planning of manipulator trajectories, leading roboticists to rely on reactive methods, such as visual servoing \cite{smith2024design}, which require real-time flower detection and pose estimation.  

Aside from the online computer vision challenges, many plants' short blooming periods significantly limit the collection of training datasets. The large number of flowers, the variation in flower shapes, and the difficulty of defining the orientation of the flowers further complicate dataset annotation. As a result of these challenges, current pollination and harvesting robots achieve a limited success rate of about 66\%, with reported ranges from 40\% to 86\% \cite{smith2024design, tang2020recognition}.
 
To address real-time flower pose estimation challenges, this work proposes digitally reconstructing plants in 3D for pose annotation and data generation and using an offline teacher model to train a more efficient online student model. Thus, our contributions to the literature are:

\begin{itemize}
    \item 3D Gaussian Splatting based training data generation, with a simple pose annotator tool.
    \item Fast and effective flower pose regression method for low computation devices.
    \item Pose refinement techniques for better localization and orientation estimation.
    \item Practical implementation using single-arm UR5 and multi-arm Stickbug~\cite{smith2024design} robots for precision pollination.
    \item Codebase, data, models, and tools are freely available at \href{https://wvu-irl.github.io/flope-irl/}{wvu-irl.github.io/flope-irl}.
\end{itemize}

The rest of the paper is outlined as follows: Section II discusses relevant works in detection, pose estimation, and training data generation; Section III details the proposed methodology; Section IV describes the experimental setup along with experimental results and discussion; Lastly, Section V concludes the work and outlines the potential future work of the study.

\section{RELATED WORKS}

\subsection{Detection and Segmentation}

Detecting plant structures, including flowers~\cite{smith2024design, cheng2020flower, singh2024deep}, tomatoes~\cite{jun2021harvest}, nodes~\cite{ci20243d}, and branches~\cite{liu2012image,schneider20233d}, is a well-known problem in agricultural robotics. Among the various approaches, recent studies have predominantly relied on YOLO~\cite{khanam2024yolov11} models for detection and segmentation, primarily due to their speed and low computational requirements. However, with the rapid advancements in foundation models~\cite{caron2021emerging, oquab2023dinov2, radford2021learning} for language and vision, new opportunities have emerged for developing robust, task-specific models for object detection and semantic segmentation. One such example is Grounding DINO~\cite{liu2025grounding}, which leverages DINO~\cite{caron2021emerging} features with text-based conditioning to accurately localize target objects within an image. Similarly, SAM~\cite{kirillov2023segment} implements a promptable model that enables flexible semantic segmentation based on input prompts. While these foundation models exhibit remarkable robustness and generalization, their high computational demands make them impractical for deployment on low-powered edge devices such as an Nvidia Jetson. By contrast, YOLO remains a more practical choice in resource-constrained settings due to its efficiency, even though it delivers lower accuracy compared to larger models~\cite{chen2019mmdetection}.

\subsection{Object Pose Estimation}

Applications like precision pollination demand flower localization in the 2D image space and estimation of accurate 6 DoF flower poses in 3D. Improper poses can be one of the most common sources of failure~\cite{kong2024towards} for such tasks. Methods for generic object pose estimation can be broadly classified into two categories: direct and indirect methods. Direct methods regress the translation and orientation components of the object directly from an RGB image. Whereas, indirect methods~\cite{sun2022onepose, he2022onepose++} first establish 2D-3D correspondences~\cite{xu2024local,shrestha2023caldiff} and recover 6-DoF pose using perspective-n-point (PNP), often in combination with the RANSAC algorithm~\cite{martin1981random} to make it robust to outliers. While indirect methods are to perform better than direct methods \cite{xu2024critical}, they require reference models like 3D point clouds or CAD models to establish correspondences with the query image. This requirement makes them less suitable for the plant structures, where the high variability in the shapes and sizes makes it impractical to compare them against a single reference model. Category-level pose estimation~\cite{fu2022category} can address these issues but comes with significant computation costs.

Keypoint-based pose estimation has been employed to estimate the pose of tomatoes~\cite{li2024ahppebot, ci20243d} and flowers~\cite{winter2018keypoint, duc2024modeling}. This approach involves detecting multiple points of interest, such as pistils, nodes, and other distinctive features on flowers and fruits, which are then used to infer the pose. Such methods fail if the target key points are not visible from the camera's viewpoint or are occluded. Point cloud alignment with the reference template object has also been used to estimate pose~\cite{jun2021harvest, kong2024towards}. However, this is not directly applicable in flower pose estimation due to flowers' more flexible and deformable nature and their variability in shapes and sizes. Additionally, the iterative point cloud alignment process can be computationally prohibitive for real-time applications, especially with many flowers in the scene.

Direct approaches like PoseNet ~\cite{kendall2015posenet} and ~\cite{shrestha2023caldiff} regress camera pose from a single image by minimizing the Euclidean norm between the estimated and ground truth camera translations and rotations. A similar strategy is employed by \cite{rambach2018learning} to regress object pose from a single image. However, the quaternions do not lie in the Euclidean manifold, but rather in the $SO(3)$ manifold. Hence, directly optimizing for Euclidean angles or quaternions will cause a sub-optimal learning path due to discontinuity and double cover~\cite{sax2009learning, geist2024learning}. To address this issue, \cite{zhou2019continuity} learns rotational features in $\mathbb{R}^6$, and projects them to $SO(3)$ using Gram-Schmidt orthonormalization, providing continuous representation space. Another approach proposed by \cite{levinson2020analysis} suggests to learn rotational features in $\mathbb{R}^9$ and project it to $3\times3$ rotational matrix in $SO(3)$ using singular value decomposition (SVD). 

% Precision pollination requires accurate 6-DoF flower pose estimation, as improper poses are a major source of failure~\cite{kong2024towards}. Generic object pose estimation methods fall into direct and indirect approaches. Direct methods regress translation and orientation from an RGB image, while indirect methods~\cite{sun2022onepose, he2022onepose++} establish 2D-3D correspondences~\cite{xu2024local,shrestha2023caldiff} and recover pose using PNP with RANSAC. Though more robust~\cite{xu2024critical}, indirect methods rely on reference models like 3D point clouds or CAD models, making them unsuitable for flowers, which exhibit high shape and size variability. Category-level pose estimation~\cite{fu2022category} mitigates this but is computationally expensive. Keypoint-based methods have been applied to tomatoes~\cite{li2024ahppebot, ci20243d} and flowers~\cite{winter2018keypoint, duc2024modeling} by detecting distinctive features like pistils and nodes, but they fail under occlusions. Point cloud alignment~\cite{jun2021harvest, kong2024towards} offers an alternative but struggles with flowers' flexible structures and is computationally prohibitive for real-time applications with multiple flowers.

% Following~\cite{levinson2020analysis}, we use the $\mathbb{R}^9+\text{SVD}$ based rotation representation and reduce the squared chordal distance~\cite{hartley2013rotational} for directly regressing flower pose given the flower image. 

\subsection{Training Data Generation}

Training an effective detection and pose estimation model requires a large amount of accurately labeled data. However, acquiring such data is labor-intensive and time-consuming. Sim2Real~\cite{junge2023Lab2FieldTO, rizzardo2020importance} techniques address this challenge by generating synthetic data, training models in simulation, and then transferring the learned behavior to real-world applications. The domain gap between simulated and real data is often substantial for vision-based tasks, leading to significant performance degradation in real-world applications. Recently, Neural Radiance Fields (NeRFs)~\cite{mildenhall2021nerf} and 3D Gaussian Splatting (3DGS)~\cite{kerbl20233d} have been shown to capture real-world scenes with high-fidelity, photorealistic image renderings, offering a potential solution to bridge this gap, and have been used in various robotic applications~\cite{irshad2024neural}. For example, \cite{ojo2023splanting} uses 3DGS technique for plant phenotyping. Additionally, SplatSim~\cite{qureshi2024splatsim} uses 3DGS rendered RGB images to learn manipulation policies.

% A common approach for representing a 3D object or scene is using a 3D point cloud, obtained using LiDAR, depth camera, or simple RGB camera using photogrammetry pipelines~\cite{schoenberger2016sfm}. While point clouds are a simple and straightforward representation, they are sparse and require additional processing for surface reconstruction. Even with surface reconstruction, the viewpoint rendering from such representations tends to lack photorealism, making them unsuitable for synthetic data generation. In contrast, Neural implicit representations, such as Neural Radiance Fields (NeRFs)~\cite{mildenhall2021nerf}, model volumetric scenes by learning a continuous function that maps 3D coordinates to color and density, achieving photorealistic novel view synthesis but at the cost of high computation and has slow rendering time. Recently, 3D Gaussian Splatting (3DGS)~\cite{kerbl20233d} has emerged as a more efficient alternative, leveraging anisotropic 3D Gaussians to render high-quality scenes in real time with improved efficiency over NeRFs. \cite{ojo2023splanting} uses 3DGS technique for plant phenotyping. We implement 3DGS reconstruction of plants, and render photorealistic images from random view points to supply training data for our models.
\section{METHODOLOGY}

\begin{figure*}
    \centering
    \includegraphics[width=\linewidth]{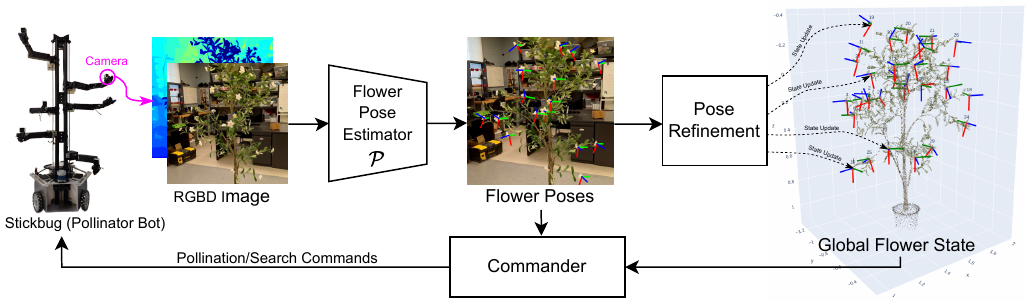}
    \caption{\textbf{Implementation Pipeline.} Stickbug captures  RGB-D images and feeds them to the Flower Pose Estimator, which predicts flower poses. These poses are refined and integrated into the global flower state. The Commander, aware of the updated state and current flower poses, decides whether to send pollination or search commands to the Stickbug.}
    \label{fig:pipeline}
    \vspace{-2mm}
\end{figure*}

\begin{figure}
    \centering
    \includegraphics[width=\linewidth]{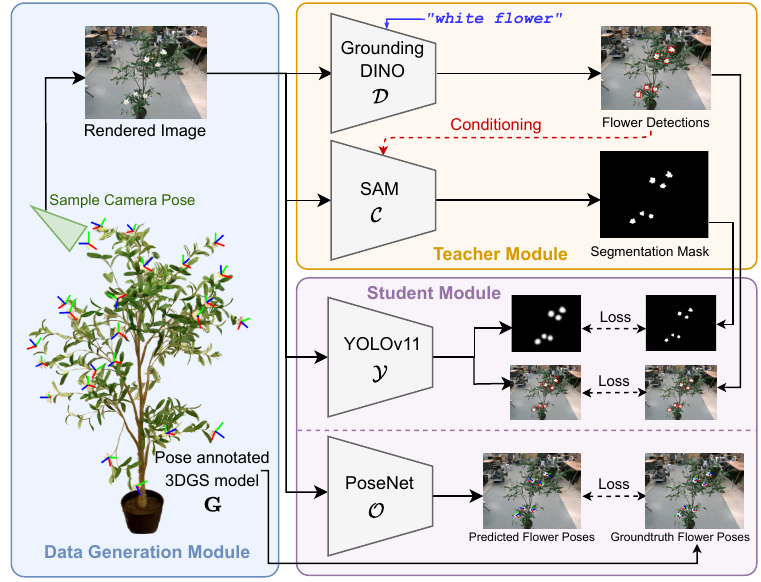}
    \caption{\textbf{Training Pipeline using Knowledge distillation.} The data generation module samples a random camera pose around the target object and renders a novel view at every training iteration. The Teacher Module generates pseudo-groundtruth labels for object bounding boxes and segmentation masks. The Student Module implements a small and fast YOLOv11~\cite{khanam2024yolov11} nano model to learn the labels as provided by the teacher module. The PoseNet model learns to estimate flowers' poses using groundtruth pose annotations.}
    \label{fig:knw_dist}
    \vspace{-3mm}
\end{figure}

This study aims to develop an accurate and efficient flower pose estimation module to guide a robotic manipulator for precision pollination. The methodology is presented in two parts: the training pipeline (Fig. \ref{fig:knw_dist}) and the implementation pipeline (Fig. \ref{fig:pipeline}). In the training pipeline, a data generation module is used to create training data, enabling knowledge distillation from a teacher module to a lightweight student module. The implementation pipeline details how the trained flower pose estimator is deployed to facilitate the pollination task.

\subsection{Data Generation}
Since flower pose estimation models rely entirely on vision, as illustrated in Fig. \ref{fig:pipeline}, where an RGBD image serves as input, we use a 3DGS~\cite{kerbl20233d} reconstruction as the plant model, leveraging its ability to generate high-fidelity, photorealistic renderings. First, a scene's video frames $\mathbf{I}$ are captured from this model. Each image in the video sequence $I_i \in \mathbb{R}^{3\times H \times W}$, is used for multi-view reconstruction using COLMAP~\cite{schoenberger2016sfm}. This provides a set of colored 3D points $p_i \in \mathbb{R}^{6}$ and corresponding camera poses $c_i$ represented by positions $x^c_i \in \mathbb{R}^3$ and quaternions $q^c_i \in \mathbb{R}^4$. $p_i$ comprises of 3D coordinate $x_i \in \mathbb{R}^{3}$ and the corresponding RGB values $r_i \in \mathbb{R}^{3}$.

The obtained 3D points $p_i$ are used to initialize the Gaussian centers of 3D Gaussian distribution $g_i$ as represented in Equation~\ref{eqn:gaussian}. $g_i$ allows for smooth, continuous surface reconstruction while maintaining high visual quality. The initialized Gaussian centers $x_i$ will get refined during the Gaussian splatting training process to generate high-fidelity visual renders.
\vspace{-3mm}
\begin{equation}
\label{eqn:gaussian}
g_i(\xi) = \sigma(\alpha_i)\exp(-\frac{1}{2}(\xi - x_i)^T\Sigma_i^{-1}(\xi - x_i))
\vspace{-2mm}
\end{equation}
where, $\xi \in \mathbb{R}^3$ represents the 3D position,  $x_i$ is the mean position of the Gaussian,  $\Sigma_i$ is the covariance matrix, $\alpha_i$ is the amplitude parameter and $\sigma(\cdot)$ is an activation function. The obtained Gaussian Splatting Model $\mathbf{G}$, is used by rasterizer function $\mathcal{R}(\mathbf{G}, c^S)$ to generate a photorealistic renders $I^G$ from any randomly sampled camera pose $c^S$.

We developed a 3D pose annotator tool that significantly simplifies annotating flower poses within the 3DGS point cloud. With this, a complete 3DGS plant model with around $20-25$ flowers can be annotated in a few minutes. The annotation process relies on human judgment to determine each flower's orientation, which introduces some degree of error, but remains a practical and reliable approach for obtaining ground-truth data. Using the annotated 3D model, flower poses are projected onto the rendered image $I^G$ from multiple viewpoints to obtain the ground-truth orientation, $R^\text{gt}$, which is then utilized for training the pose estimation module.

% where, $\xi \in \mathbb{R}^3$ represents the 3D position,  $x_i$ is the mean position of the Gaussian,  $\Sigma_i$ is the covariance matrix, $\alpha_i$ is the amplitude parameter and $\sigma(\cdot)$ is an activation function. Each Gaussian blob $G_i$ in a trained Gaussian splatting model $\mathbf{G}$ comprises of 3D Gaussian center $x_i$, normals $n_i \in \mathbb{R}^3$, color values $r_i$, spherical harmonics $sh_i \in \mathbb{R}^{45}$, and opacity values $o_i \in \mathbb{R}^1$. A Gaussian rasterizer function $\mathcal{R}(\mathbf{G}, c^S)$ is defined to generate a photorealistic image $I^G$ given a set of 3D Gaussians $\mathbf{G}$ and a randomly sampled camera pose $c^S$.

\subsection{Flower detection and segmentation}

A knowledge distillation strategy is implemented to transfer knowledge from a teacher model to a smaller and faster student model, as shown in Fig. \ref{fig:knw_dist}. For each rasterized image $I^G$, pseudo-groundtruth flower detection bounding boxes $b \in \mathbb{R}^{F \times 4}$ are generated using the Grounding DINO~\cite{liu2025grounding} model $\mathcal{D}(I^G, t)$, conditioned upon a text prompt $t$=\textit{"white flower"}. Here, $F$ denotes the number of flowers detected in $I^G$, and the bounding boxes $b$ are used as input to the Segment Anything Model (SAM)~\cite{kirillov2023segment} $\mathcal{C}(I^G, b)$, to generate flower segmentation mask $m\in\mathbb{R}^{H \times W}$ for the given image $I^G$. Finally, $b$ and $m$ are used to train the YOLOv11~\cite{khanam2024yolov11} student model $\mathcal{Y}(I)$ using the training pipeline provided by Ultralytics~\cite{ultralytics}. During inference, only $\mathcal{Y}$ is used to detect flowers and their segmentation mask.

\subsection{Flower Pose Estimation}

 PoseNet~\cite{kendall2015posenet} is utilized as the orientation estimation model $\mathcal{O}(I[b])$ to directly regress the rotational features $\rho \in \mathbb{R}^9$. The rotational features $\rho$ are reshaped to a $3 \times 3$ matrix and projected to a rotational flower representation $R^f \in SO(3)$ using Singular Value Decomposition. Here, $[.]$ denotes a cropping operation, where bounding boxes $b$ crop out patches from $I$, which are fed into $\mathcal{O}$. The choice of using $\mathbb{R}^9+\text{SVD}$ based rotational representation is vital for the learning process of the $\mathcal{O}$. This provides a continuous representation space without ambiguity (e.g., double cover), ensuring stable and nonfluctuating gradients during training. The model $\mathcal{O}$ is trained by minimizing $||R^f-R^{\text{gt}}||$, where $||.||$ represents the L2 norm. To account for the radial symmetry of the flowers, yaw angles are nullified before computing the loss. The camera intrinsic parameters $K$ and depth image $D$ are then used to map the 2D pixel coordinates $u$ to flower 3D coordinates $x^f$ as shown in Equation \ref{eqn:uplift}.
\vspace{-2mm}

\begin{equation}
x^f = D[u]\frac{K^{-1}\tilde{u}}{||K^{-1}\tilde{u}||}
\label{eqn:uplift}
\end{equation}

where, $\tilde{u}$ represent homogeneous form of $u$, $D[u]$ is depth values extracted at the pixel coordinates $u$. Note that $D[u]$ is the depth along the ray (not along the z-axis of the camera coordinate frame), necessitating the normalization factor.

Finally, the flower's 3D coordinates are transformed to the world coordinate frame using the camera pose matrix $C$. This is necessary to ensure a consistent flower position across different viewpoints. This completes our Flower Pose Estimator $\mathcal{P}(I,D)$ module, which estimates 6 DoF flower poses $p^f=\{R^f,x^f\}$ using $I$ and $D$.

% \begin{equation}
% x^{f,W} = C.x^f
% \end{equation}

\subsection{Pose Refinement}
The pose predicted by $\mathcal{P}$ is inherently noisy due to measurement errors from the depth camera and inaccuracies in orientation estimation by $\mathcal{O}$. An Extended Kalman filter is applied to the position $x^f$ and orientation $R^f$ of the flower to obtain smoother estimates. 

Since $x^f$ belongs to $\mathbb{R}^3$, filtering is simple with smooth interpolation between the predicted and measured state. However, orientation in $R^f \in SO(3)$ suffers from discontinuities and singularities when represented using Euler angles or quaternions, making direct filtering unstable; thus, it is converted to the $\mathbb{R}^9 + \text{SVD}$ representation, with smooth interpolation \cite{levinson2020analysis}. Specifically, the rotational state and measurements are represented in $\mathbb{R}^9$ by flattening the $3 \times 3$ rotation matrix $R^f$. The Kalman update step is then performed in this space. After each update, the resulting state is projected back onto $SO(3)$ using SVD before proceeding to the next Kalman update step. This ensures that the rotational state remains within $SO(3)$ while being smoothly updated with new rotational measurements.

The global state of all flowers in the scene is maintained and refined with each new measurement using the Kalman update. For each detected flower, its Euclidean distance to all flowers in the global state is computed, and it is assigned to the closest match if within a predefined threshold. The assigned measurement then updates the corresponding flower's pose in the global state.

\subsection{Commander}

The Commander, as shown in Fig. \ref{fig:pipeline}, serves as the interface between the pollinator’s memory, which maintains the estimated global state of flowers, and the physical hardware, dynamically switching between searching and pollination modes. In searching mode, the manipulator explores the workspace in random directions, collecting measurements from different viewpoints to refine flower pose estimates. Once a flower’s estimated pose reaches a sufficient confidence level with low variance, the system transitions to pollination mode, which consists of two phases. The first phase, rough localization, moves the manipulator near the estimated flower pose but stops a few centimeters short to prevent misalignment to account for potential inaccuracies, ensuring the flower remains within the camera’s field of view. The second phase, visual servoing, discards the global estimate and fine-tunes the manipulator’s position based on real-time visual feedback. The difference between the end effector’s current position and the target flower’s pose is used to compute a movement gradient, iteratively adjusting the joint angles until the pollinator tip~\cite{smith2024design,ohi2018design} precisely aligns with the flower. Once aligned, the pollination sequence is triggered, allowing the tip to rub the flower and transfer pollen. After pollination, the Commander switches back to searching mode, maintaining a record of pollinated and unpollinated flowers to systematically continue the process until all flowers are pollinated.

\subsection{Practical implementations}
The proposed methods are deployed on two robotic platforms: a single-arm UR5 and a multi-arm Stickbug, demonstrating pollination accuracy on both. The UR5, being an off-the-shelf manipulator, is robust, easy to use, and very precise, making it well-suited for accurate flower localization. Stickbug is a custom-built robot specifically designed for precision pollination. With its six-arm configuration, each equipped with a camera, it can simultaneously capture flower data from six different viewpoints, enabling continuous updates to the global flower state and enhancing localization accuracy.

\section{EXPERIMENTS}

We begin by single shot pose estimation using $\mathcal{P}$, followed by the pose refinement techniques and finally experiments on the pollination performation.

\subsection{Flower Pose Estimation Accuracy}

% ---------------------------- Pose Metrics Table ----------------------------
\begin{table}[]
\caption{Flower Detection and Pose Estimation Accuracy}
\begin{tabular}{p{0.5mm}l|lll}
  &                         & \textbf{Teacher} & \textbf{Student} & \textbf{FloPE} \\
\hline
\multirow{3}{*}{\rotatebox{90}{\textbf{2D}}} & Detection Error (pixels) $\downarrow$        & 8.87 & 8.97 &  {\scriptsize NA$^{*}$} \\
& Detection Success Rate (\%) $\uparrow$     & 93.47 & 93.01 & {\scriptsize NA}\\
\hline
\multirow{3}{*}{\rotatebox{90}{\textbf{3D}}} & Transitional Error (cm) $\downarrow$         & 3.02 & 3.03 & 0.6 \\
&Rotational Error (degrees) $\downarrow$      & 29.78 & 29.88 & 19.14 \\
&Success Rate (\%) $\uparrow$               & 65.43 & 64.90 & 77.78 \\
\hline
\multirow{3}{*}{\rotatebox{90}{\textbf{Speed}}} & Speed in RTX3090Ti (FPS) $\uparrow$  & 1.64 & 15.5 & 15.5 \\
& Speed in Jetson (FPS) $\uparrow$  & {\scriptsize NA} & 3.45 & 3.45 \\
& Model Size (GB) $\downarrow$ & 3.86 & 0.77 & 0.77 \\
\end{tabular}
{\begin{flushright}\scriptsize {\scriptsize $^{*}$NA} = Not Applicable\end{flushright}}
\label{tab:metrics}
\vspace{-2mm}
\end{table}
% ---------------------------------------------------------------------------

% ------------------- pose metrics table old -----------------------------
% \begin{table}[]
% \caption{Flower Detection and Pose Estimation Accuracy}
% \begin{tabular}{p{0.5mm}l|p{7mm}lp{5mm}}
%   &                         & \textbf{Teacher} & \textbf{Student} & \textbf{Distil} \\
% \hline
% \multirow{3}{*}{\rotatebox{90}{\textbf{2D}}} & Detection Error (pixels) $\downarrow$        & 8.87 & 8.97 & 2.58 \\
% & Detection Success Rate (\%) $\uparrow$     & 93.47 & 93.01 & {\scriptsize NA$^{*}$}\\
% & Masking Accuracy (DICE) $\uparrow$         & {\scriptsize NA} & {\scriptsize NA} & 0.832 \\
% \hline
% \multirow{3}{*}{\rotatebox{90}{\textbf{3D}}} & Transitional Error (cm) $\downarrow$         & 3.02 & 3.03 \textbf{(0.6)} & 0.76 \\
% &Rotational Error (degrees) $\downarrow$      & 29.78 & 29.88 \textbf{(19.14)} & 22.81 \\
% &Success Rate (\%) $\uparrow$               & 65.43 & 64.90 \textbf{(77.78)} & {\scriptsize NA}\\
% \hline
% \multirow{3}{*}{\rotatebox{90}{\textbf{Speed}}} & Speed in RTX3090Ti (FPS) $\uparrow$  & 1.64 & 15.5 & {\scriptsize NA} \\
% & Speed in Jetson (FPS) $\uparrow$  & {\scriptsize NA} & 3.45 & {\scriptsize NA} \\
% & Model Size (GB) $\downarrow$ & 3.86 & 0.77 & {\scriptsize NA} \\
% \end{tabular}
% {\begin{flushright}\textbf{Bold} values represent Student Model after Pose Refinement step\\\scriptsize {\scriptsize $^{*}$NA} = Not Applicable\end{flushright}}
% \label{tab:metrics}
% \vspace{-5mm}
% \end{table}
% ---------------------------------------------------------------------------

\begin{figure}
    \centering
    \includegraphics[width=\linewidth]{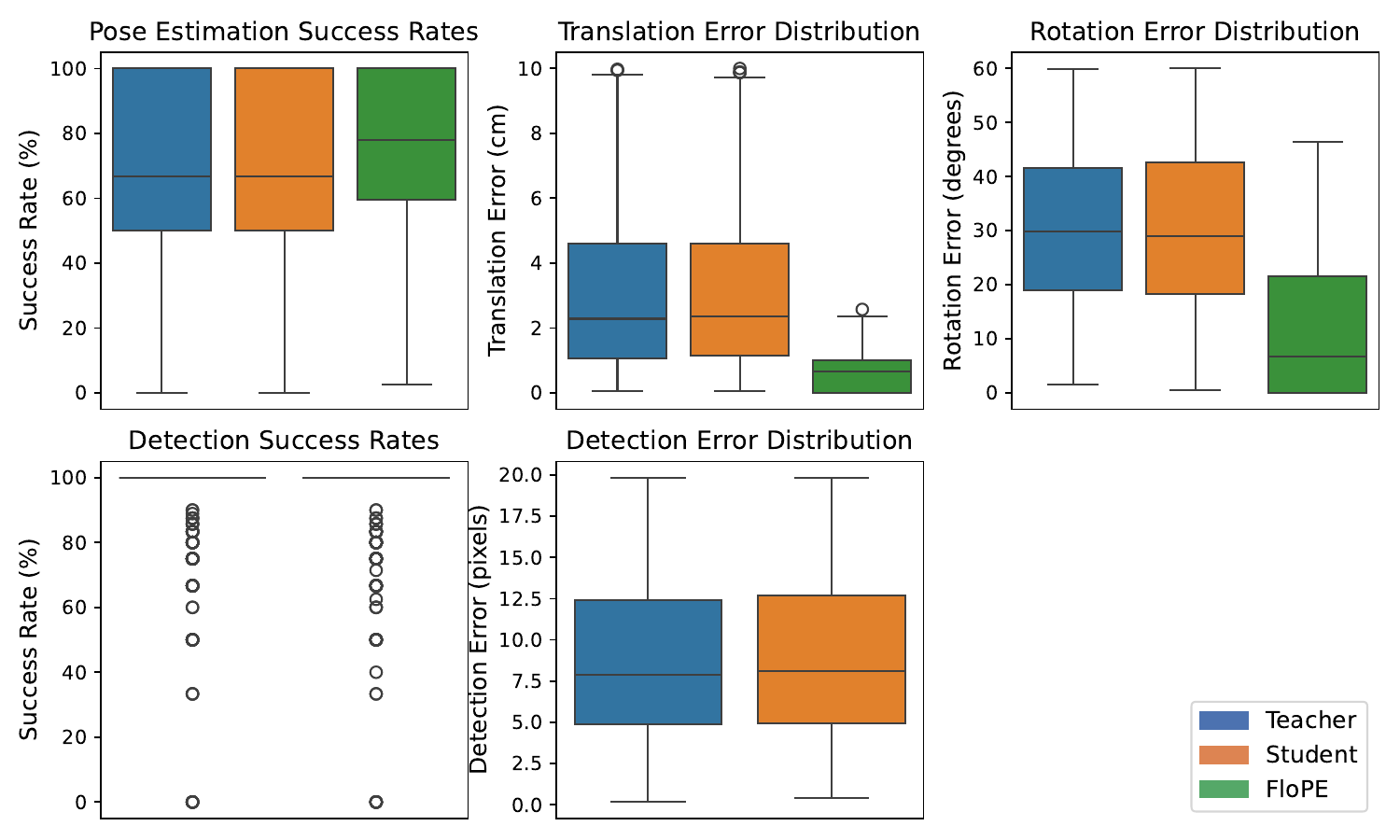}
    \caption{\textbf{Metrics Visualization.} Accuracy and error metrics from teacher, student and our final implementation \textbf{FloPE} are visualized. \textbf{FloPE} represents the student model after incorporating the pose refinement step. Note that the teacher and student models provide single-shot estimates, whereas \textbf{FloPE} generates a multi-shot estimate by filtering the student measurements.}
    \label{fig:metrics_vis}
    \vspace{-3mm}
\end{figure}

\begin{figure}
    \centering
    \includegraphics[width=\linewidth]{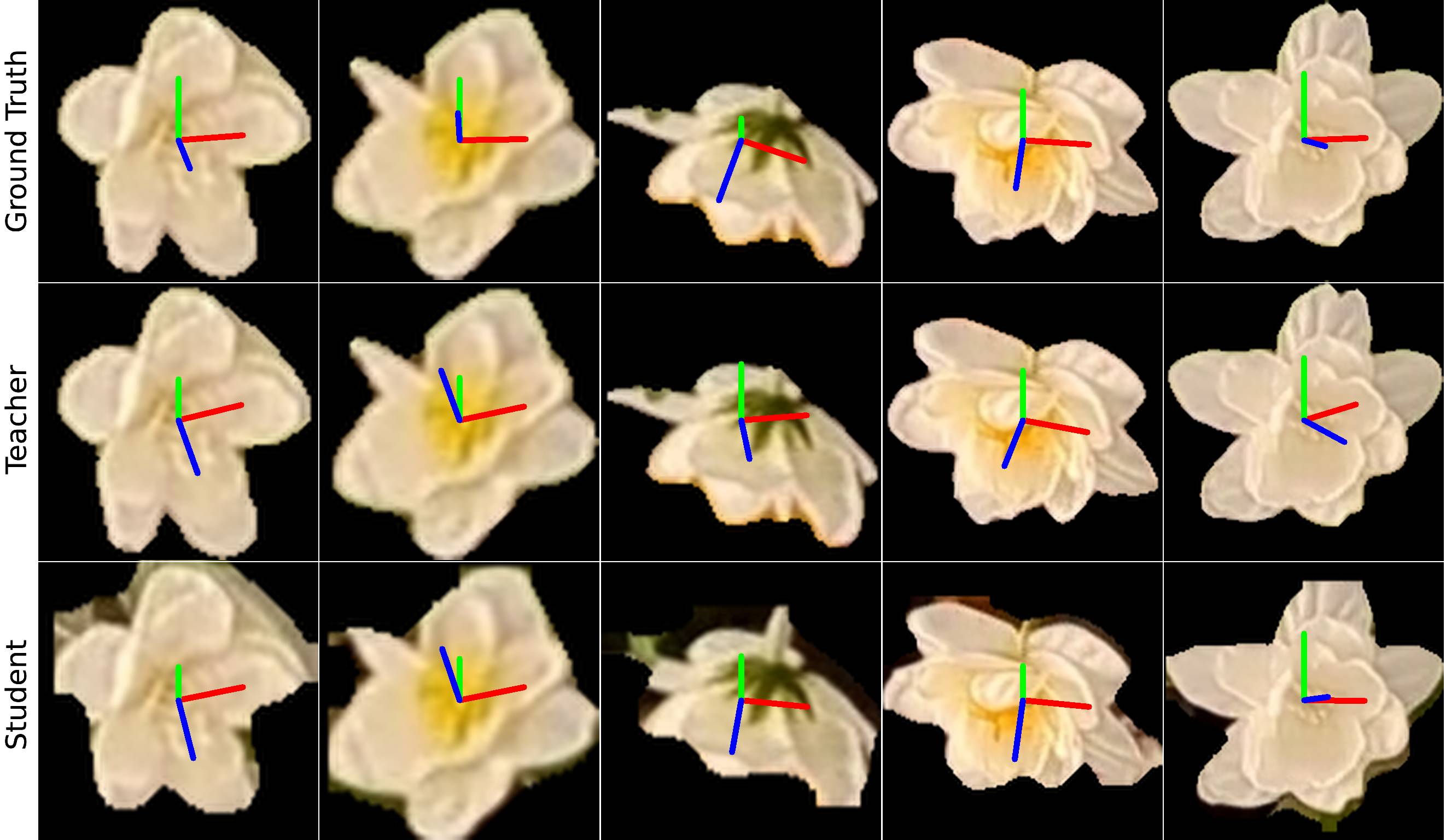}
    \caption{\textbf{Flower Poses Visualization.} Ground-truth flower poses and the estimated poses from the teacher and student models are visualized for five randomly selected flower samples. Flowers are masked using the corresponding segmentation masks provided by each model. {\color{blue}Blue} Z-axis indicates the flower's facing direction.}
    \label{fig:patches}
    \vspace{-3mm}
\end{figure}

Evaluating predicted flower poses on a real plant is challenging due to the absence of reliable ground truth data. To address this, we generate a test dataset using a separate 3DGS plant model with accurately labeled flower pose annotations, with a different flower distribution than the training data. The evaluation metrics, summarized in Table \ref{tab:metrics} and visualized in Fig. \ref{fig:metrics_vis}, fall into three key categories: \textbf{2D metrics}, which assess the accuracy of flower detection in the image plane; \textbf{3D metrics}, which evaluate the precision of 6-degree-of-freedom (6-DoF) flower pose estimation; and \textbf{speed metrics}, which measure the computational performance of the models across different hardware platforms. Here, rotational error represents angle between direction of flower (i.e. blue z-axis in Figure~\ref{fig:patches}) in predicted and ground-truth flower poses. We report these metrics for both the teacher and student models, as well as \textbf{FloPE}, our complete implementation.

Given an input image size of 1280x720, both models achieve a low detection error of approximately 8 pixels. We define a detection as successful if the error is within 20 pixels, which is used to compute the detection success rate. For pose estimation, we achieved approximately $3$ cm, $30\degree$ error before the refinement step. Since the detection error is minimal, most of the translational error arises from inaccuracies in depth estimation provided by the RealSense D405 camera. We found that depth measurements of this camera become unreliable outside its optimal operational range of 7–50 cm. The higher rotational error is likely due to ambiguities in the flower’s visual appearance from certain viewpoints. Accurately determining a flower’s precise orientation within a few degrees is inherently challenging, even for human observers when relying on a single viewpoint. To improve annotation accuracy, our annotation tool provides the ability to rotate and scale the model as needed, reducing pone annotation error.

A pose estimation is considered successful if the translational error is within 8 cm and the rotational error is within 60°. While these thresholds may seem large for precision pollination tasks, they provide a reasonable criterion for the subsequent pose fine-tuning step. This approach allows a greater number of data points to contribute to the filtering process, ultimately refining the pose estimate, as demonstrated by the FloPE metrics. After the filtering step, we achieve a translational error below 1 cm and a rotational error below $20^\circ$, with a success rate of 77.78\%. This level of accuracy enables precise localization of the flower and its orientation, facilitating successful pollination.

Looking into the speed metric, the student model achieves nearly a $10\times$ speed boost compared to the teacher model while being $5\times$ smaller in size, with only a minimal decrease in accuracy. It runs at 3.45 FPS on an NVIDIA Jetson Nano. Due to memory and processing limitations, we could not evaluate the teacher model's performance on this device. Since the filtering process in \textbf{FloPE} introduces negligible computational overhead, it operates at the same speed as the student model.

Moreover, we evaluate the distillation error for masking by calculating the DICE score~\cite{azad2023loss} between the flower segmentation masks generated by the teacher and student models. The obtained DICE score of 0.832 indicates a good agreement between the two models. The last two rows of Fig. \ref{fig:patches} visually compare the masked flowers for both models.

\subsection{Pollination Performance}
% ---------------------------- Pollination Metrics Table ----------------------------
\begin{table}[]
\caption{Pollination Performance Metrics}
\begin{tabular}{l|c|cc}
& \textbf{Previous Work} & \multicolumn{2}{c}{\textbf{FloPE(Ours)}}  \\ 
\textbf{Pollination Metrics} & \textbf{StickBug~\cite{smith2024design}} & \textbf{StickBug}  &  \textbf{UR5 Arm} \\ 
\hline
Attempt Rate (\%) $\uparrow$   & 83.33     & 73.33     & 90.0 \\ 
Success Rate (\%) $\uparrow$   & 49.00     & 61.36     & 78.75               
\end{tabular}
\label{tab:pollination_metrics}
\vspace{-3mm}
\end{table}
% ---------------------------------------------------------------------------

Our experimental setup to evaluate pollination performance consists of a robotic manipulator with an eye-in-hand configuration and a pollinator tool in its end effector. We design two different experimental setups: one using a single-arm UR5 manipulator and another using a multi-arm StickBug~\cite{smith2024design} robot. To ensure a feasible experimental setup, we use artificial bramble flowers that closely resemble real ones. This choice is necessary since real bramble flowers are unavailable during the winter months, making them impractical for our project timeline.

A successful pollination is achieved through single-shot pose estimation, followed by pose fine-tuning and visual servoing. We evaluate performance using two key metrics: pollination attempt rate, which quantifies the number of attempts made to pollinate flowers within the robot manipulator’s reach, and pollination success rate, which measures the proportion of successful pollination attempts. A pollination attempt is considered successful if the pollinator tip makes contact with the pistil of the flower. Each experiment is repeated five times, and the average results are reported in Table \ref{tab:pollination_metrics}. The proposed method outperformed the previous work with the stickbug robot for pollination success rate but had reduced pollination attempts. This likely occurred due to the previous work attempting to pollinate false positive flowers and also completely ignoring the flower orientation \cite{smith2024design}. Moreover, the UR5 arm had a very high pollination attempt and success rate. 

Since the StickBug arm is a custom-built manipulator with fewer degrees of freedom than the UR5, its lower success rate is expected. Most failures with the StickBug occurred due to ineffective manipulator control, where other parts of the arm inadvertently displaced the flower before the pollinator tip could make contact, resulting in unsuccessful pollination attempts.
\section{CONCLUSION AND FUTURE WORK}

This work addresses the critical problem in agricultural robotics of accurately estimating the pose of plant structures while operating under limited onboard computational resources. This approach can potentially enable a wide range of applications that require precise pose estimation of plant structures. The method was demonstrated on flowers in pollination tasks, but the same strategy can be extended to estimate the pose of other plant structures. Additionally, our approach of training a lightweight model using synthetic data and a knowledge distillation strategy presents a promising direction for agricultural robotics, where the scarcity of high-quality training data is a significant challenge.

The primary failure cases in our pose estimation arise when the estimator $\mathcal{P}$ produces two or more equally likely pose predictions, which disrupts the filtering mechanism. One possible solution is leveraging category-level pose estimation or feature-matching techniques. Exploring knowledge distillation from such models could enable the development of a lightweight yet robust pose estimator. Moreover, our current flower search strategy relies on a random approach, which may result in missed flowers. Implementing active vision techniques could allow the system to search systematically for unpollinated flowers, improving overall efficiency in pollination tasks. Future work will address these challenges and refine the method for broader applications in agricultural automation.

% \addtolength{\textheight}{-12cm}

\section*{ACKNOWLEDGMENT}
Computational resources were provided by the WVU Research Computing Dolly Sods HPC cluster, which is funded in part by NSF OAC-2117575. We extend our gratitude to Dr. Nicole Waterland and her students for providing access to the greenhouse facility.

\bibliographystyle{IEEEtran}
\bibliography{egbib}

% Appendix here

\end{document}